# Splitting Convolutional Neural Network Structures for Efficient Inference

*Emad MalekHosseini[1], Mohsen Hajabdollahi[1], Nader Karimi[1], Shadrokh Samavi[1,2], Shahram Shirani[2]*

[1]Department of Electrical and Computer Engineering, Isfahan University of Technology, Isfahan, 84156-83111 Iran

[2]Department of Electrical and Computer Engineering, McMaster University, Hamilton, ON L8S 4L8, Canada

**ABSTRACT**

For convolutional neural networks (CNNs) that have a large volume of input data, memory management becomes a major concern. Memory cost reduction can be an effective way to deal with these problems that can be realized through different techniques such as feature map pruning, input data splitting, etc. Among various methods existing in this area of research, splitting the network structure is an interesting research field, and there are a few works done in this area. In this study, the problem of reducing memory utilization using network structure splitting is addressed. A new technique is proposed to split the network structure into small parts that consume lower memory than the original network. The split parts can be processed almost separately, which provides an essential role for better memory management. The split approach has been tested on two well-known network structures of VGG16 and ResNet18 for the classification of CIFAR10 images. Simulation results show that the splitting method reduces both the number of computational operations as well as the amount of memory consumption.

## 1. INTRODUCTION

Convolutional neural networks (CNNs) have a large number of parameters and many elements that require memory. In a practical situation, when large input data are employed, the memory usage and memory management can be a bottleneck to reach an appropriate performance. Extensive input data generate correspondingly large feature maps that consume extensive memory. On the other hand, in the CNN processing, all of the input feature maps should be processed simultaneously, and all of them are required at the same time to have efficient processing [1].

There are several methods to handle the problems existing in the CNN memory management. In [2]–[4] memory reduction is addressed by using feature map pruning. Feature maps can be removed by removing their corresponded filters. In [5] a method for feature map pruning in medical applications is proposed. Pruning techniques yield an irregular structure in which the ease of implementation may not be consistent with the pruning ratio [6]. Also, pruning techniques are restricted to the network structure, and a structure itself may have a problem with large memory consumption.

A common way to deal with the problem of large feature maps is tiling, which necessitates a lot of data transactions. In [7], the reduction of input size is considered to achieve a network with relatively small feature maps. The input image is split into patches during training, and with a small amount of padding, utilization of small-sized feature maps becomes possible. This input splitting is tested in the case of medical images; however, in the case of natural images, a general solution hasn't been proposed.

Memory reduction can be realized by designing an efficient network structure employing small feature maps through manual or automatic methods [8], [9]. In [10], an efficient network structure including a decision block and multiple branches is proposed. In their structure, each branch is specialized for different inputs. The proposed branches can be designed dynamically in such a way that a simple structure results. In [11] a flexible deep neural network is proposed in which a new stage of the network is processed if the results are not sufficient. In this way, on-demand processing can be achieved. In [12], [13], a tree CNN is proposed, in which a classifier in the root of the structure is responsible for the coarse classification, while several classifiers are dedicated to fine classification. In [14]–[16], a multipath structure is proposed to reach a suitable detection performance; however, the problem of reducing memory usage is not considered. In [17] a progressive structure incrementally boosts the network structure's performance. In this work, modifying the depth of the network is addressed but modifying its width is not. In [18] and [19] a primary part is considered for the global feature extraction and several branches are responsible for specialization.

In [20] the problem of network splitting is addressed. The network structure is divided in such a way that several slim sub-networks are created. Although the issue of network splitting is studied, the network structure is tested only on a face detection application. In the case of general classification tasks, the network splitting may not have a straightforward solution, and a complete splitting may not be possible. In this paper, the problem of network structure splitting for achieving several slim networks is investigated. Several network parts are considered and the network is split into slim ones. The splitting process is continued until no significant performance loss is observed. After splitting each part, a fusion layer is responsible for combining the results of the split networks. Fusion is conducted at the end of the split part to preserve network performance. This fusion is conducted after pooling operations and combines the results of the pooling layers. Using the proposed method, all parts of the network can be split as much as possible and through that, fewer hardware resources can be utilized. The remaining part of this study is organized as follows. In Section 2, CNN model splitting is explained and the proposed method is presented, and the complexity of different structures is analyzed. In Section 3, experimental results are described, and finally, Section 4 is dedicated to the concluding remarks.

## 2. PROPOSED CNN MODEL SPLITTING

### 2.1. CNN model splitting

CNNs have a large number of feature maps that should be processed simultaneously. Splitting can be very useful for having a small number of feature maps in each layer. In Fig. 1, a two layers CNN model and its split version are illustrated. The original model is split into two parts, which work entirely separately. Hence this splitting is named as an "ideal splitting" in which there exist no connections between the split parts. As illustrated in Fig. 1, in the ideal splitting, the number of feature maps which should be processed simultaneously can be reduced significantly. Also, the number of computational operations can be reduced. However, an ideal splitting cannot be possible in the case of all networks. Recent studies tried to split the network structure as much as possible. In this regard, in [21], a part of the network is considered shared without any splitting. Split parts are dedicated to the network structures which are connected to the shared base. Although there are research works on the problem of network splitting, an automatic solution for splitting the current network structures can be very beneficial.

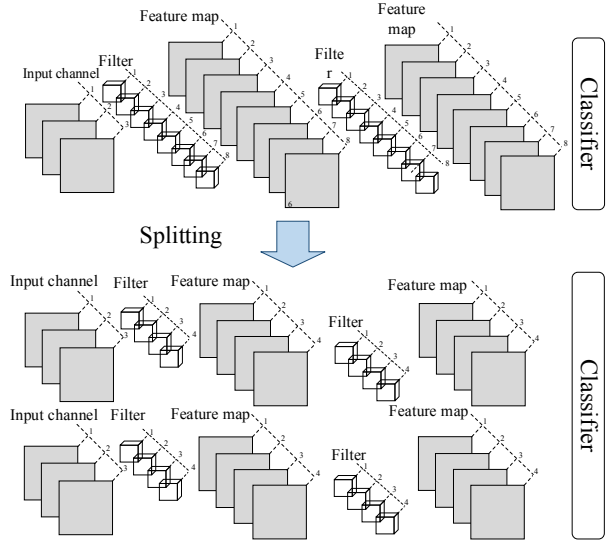

Fig. 1 Ideal model splitting in CNN

### 2.2. Proposed method

In Fig. 2, the proposed split model for CNN is depicted. Two parts of a sample CNN are illustrated. $Conv\_Block\ i$ could be one or multiple convolutional layers. The parameters $K_1$ and $K_2$ are the splitting factors in the first and second blocks, respectively. $Conv\_Block\ i\ /\ K_i$ is the split of $Conv\_Block\ i$ with parameter $K_i$. As illustrated in Fig. 2, a fusion part is designed after the splitting to preserve global information and provide a suitable input size to the next split block. Fusion block is shown in Fig. 3, which constitutes three steps including pooling operation, concatenation, and 1×1 convolution. The model illustrated in Fig. 2 is the outcome of the algorithm shown in Fig. 4 as pseudo-code. At first, a baseline network is trained to reach its final accuracy. After that, some blocks are considered for splitting. For better consistency and more reduction in the number of feature maps, the fusion block can be located on the location of pooling operations in the original network. This is illustrated in line 3 of Fig. 4. Therefore the number of split blocks could be the same as the number of pooling operations. The first feature maps in the CNNs have a larger size than the final ones. Therefore, the splitting process starts from the first layer by a splitting factor of two, meaning that the first block is divided by two. After that, the model is fine-tuned in a few epochs, and its accuracy is evaluated. In each block, splitting is continued until a higher than a given threshold of accuracy is observed. As illustrated in line 5 of the algorithm of Fig. 4, the splitting process is conducted block by block for all of the network blocks. In this way, each block can be split based on its capacity for splitting. Based on the proposed method, some layers may not be split at all due to their importance and deep feature map dependencies.

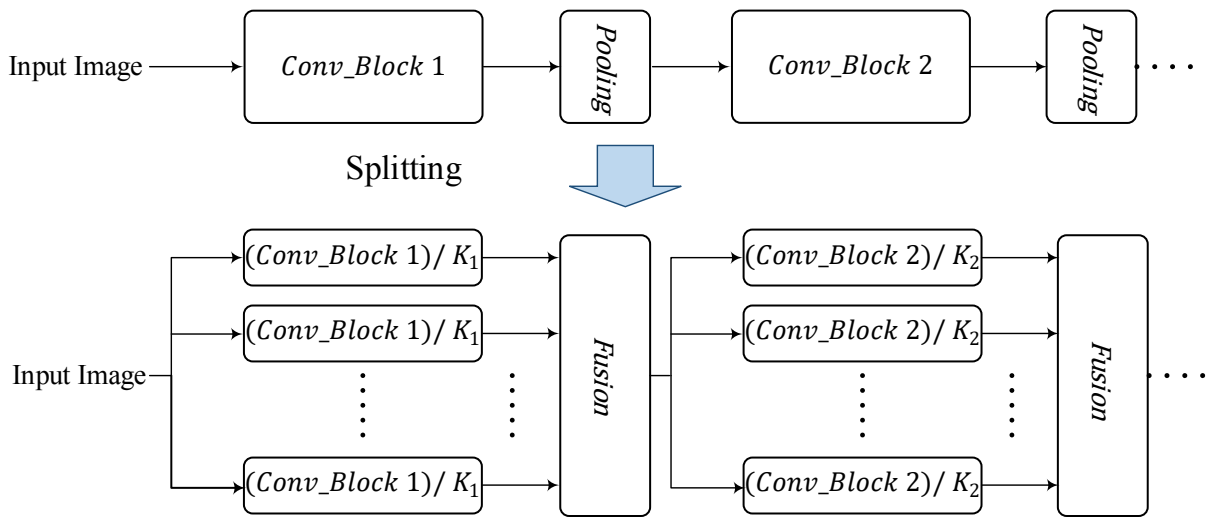

Fig 2. Proposed splitting model for CNN

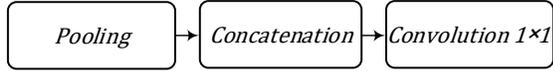

Fig. 3. Fusion block.

### 2.3. Computational and memory analysis of the proposed splitting method

CNNs suffer from needing too many parameters, and this problem is aggravated when the input data have a large size. Also, there are a lot of computational operations and many parameters in a convolutional neural network. Splitting can be used as an efficient way to alleviate problems of the model complexity existing in a typical CNN. Consider a two-layer CNN with $L_0$ input channels, $L_1$ and $L_2$ convolutional layers in the first and the second layers with 3×3 kernels, respectively. Also, assume the splitting factors $k_1$ and $k_2$ for the first and second layers, respectively. The number of parameters in the original version named as $Params_{org}$ can be calculated as equation (1). Also, the number of parameters in the split model named as $Params_{split}$ can be computed based on equation (2).

$$Params_{org} = (L_0 \times L_1 + L_1 \times L_2) \times 3^2 \quad (1)$$

$$Params_{split} = \left(\left(L_0 \times \frac{L_1}{k_1}\right) \times k_1 + \left(\frac{L_1}{k_1} \times \frac{L_2}{k_2}\right) \times k_2 \right) \times 3^2 \quad (2)$$

From equation (1, 2), it is possible to visualize the number of operations as Fig. 5. Based on different size of splitting $k_1$ and $k_2$, the number of parameters is compared with a model, which is the first two layers of VGG16. It can be observed that when the proposed algorithm couldn't split any block of the model, the outcome is the same number of parameters. By increasing the split size, less number of parameters are used. Another advantage of the splitting method is removing the tight dependencies existing between all the layer's feature maps. For example, in the second layer of VGG16, 64 input feature maps should be processed simultaneously. Although there are methods proposed to deal with the problem of the large size feature maps such as tiling, they require a lot of data movements and memory transactions. Using the proposed splitting method, the number of feature maps which should be processed simultaneously can be reduced with parameters $k_i$. Splitting reduces the number of memory operations as the most expensive operations [22].

## 3. EXPERIMENTAL RESULTS

For evaluation of the splitting method, VGG16 and ResNet18 structures are selected as the model, and the CIFAR10 image dataset [23] is used to train and test the original and the split models for the classification task. All of the codes are developed using Python on the TensorFlow framework and, a PC equipped with an Nvidia GPU 1080 Ti with 11GB DRAM is used for running the simulations.

VGG16 and ResNet18 structures are split based on their pooling locations, which means that there are five fusion blocks. As stated before, the splitting algorithm begins from the first layer of a model. In Table 1, the trend of splitting from the first block to the last one is illustrated for the ResNet18 model. Splitting factors of 1, 2, 4, 6, and 8 are tested, and the selected split factors are highlighted. The threshold for the allowed accuracy degradation is set to 0.5. It can be observed that the effect of splitting on the final model accuracy can be different in different stages. In some layers, splitting improves the accuracy resulted from the original network.

---

| 1 | Input: From network $F(X,W)$ derive the network $F'(X,W')$ with less feature map dependency |
|---|---|
| 2 | $P$: Number of pooling layers in $F(X,W)$ |
| 3 | $B$: Number of blocks in $F'(X,W')$, usually $B = P$ |
| 4 | $\{K\}_{i=1}^{B}$: The splitting factors for block $i$ in $F'(X,W')$ e.g. $K_i \in \{2,4,6,8\}$ |
| 5 | For $i$ in $1:B$ |
| 6 |    Attach block $i$ (Start in case of block $i = 1$) |
| 7 |    Train the current $F'(X,W')$ with block $i$ having a splitting factor of 1 to obtain the baseline accuracy |
| 8 |    For $j \in K_i$ |
| 9 |      Split block $i$ with the splitting factor $j$ |
| 10 |      Apply fusion and Train the split network |
| 11 |      Compare the obtained accuracy with the baseline accuracy to obtain $\Delta a$ |
| 12 |      If $\Delta a < Threshold\ T$ |
| 13 |         Continue the loop |
| 14 |      Else |
| 15 |         Exit the loop |
| 16 |    Fix the current split architecture of $F'(X,W')$ up to block $i$ |

Fig. 4. Splitting algorithm

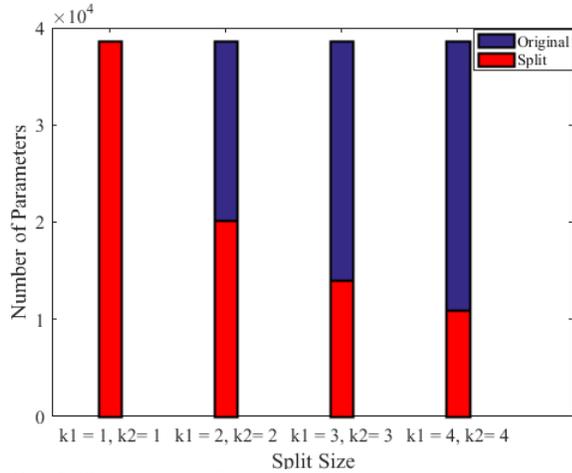

Fig. 5. The number of parameters in the original model and split one. Model is considered the first two layers of VGG16.

By setting a different threshold for the allowed accuracy drop in the proposed algorithm, different structures can be produced. By applying the proposed splitting method on the VGG16 structure for two thresholds, two different split networks are obtained. These structures are named as *proposed_I* network with splitting factors 8, 2, 2, 4, 4, and the *proposed_II* network with splitting factors 8, 2, 2, 2, 4. Two other approaches for network splitting besides of the proposed method are also tested. One of them is an approach (named as naïve split network) in which a network is split into several slim networks that are connected at the end of their structures [20]. The second one is dedicating a shared part to the first layers and splits the other parts of the network (named as the shared split network) [13]. In the shared split network, the first three layers of the VGG16 network are dedicated as the shared part and after that, the structure of the network is split with splitting factor four. The accuracy, as well as the number of parameters in the networks mentioned above, are represented in Table 2. It can be observed that by using the proposed splitting algorithm, all parts of the VGG16 structure could be split without significant accuracy degradation. The naïve approach for splitting yields a simple structure but its accuracy degradation can be regarded as high. By the splitting method introduced in [13], better accuracy is resulted. Based on the results illustrated in Table 2, we can compare the proposed and the shared splitting networks. Better accuracy is resulted in the proposed_I network with a slightly more complex structure. Splitting in the shared splitting method is not conducted in the three first layers of the VGG structure. This means that large feature maps in the first layer of the network is not split. In our splitting methods, all of the layers are split, while in the first layer, more splitting is performed. In the first layer of a CNN, there are larger feature maps as compared to the last layers. Hence, the primary layers require heavy memory operations. These massive memory operations, which are caused by large feature maps, could be reduced with the proposed splitting method.

Table 1. Trend of splitting in ResNet18

| Splitting Factor / Block | 1 | 2 | 4 | 6 | 8 |
|---|---|---|---|---|---|
| Block1 | 93.57 | 93.64 | 93.55 | 93.41 | **93.46** |
| Block2 | 93.46 | 93.36 | **93.38** | 93.03 | 93.03 |
| Block3 | 93.38 | **93.06** | 92.78 | 92.95 | 92.75 |
| Block4 | 93.06 | **93.21** | 92.89 | 92.71 | 92.34 |
| Block5 | 93.21 | 93.16 | 93.20 | **93.45** | 93.14 |

Table 2. The number of parameters in different splitting approaches

| Networks | Accuracy | # of convolutional weights (in thousands) |
|---|---|---|
| Original | 92.12 | 1634 |
| Naïve Splitting [20] | 89.81 | 408 |
| Shared Splitting [13] | 91.31 | 430 |
| Proposed_I | 91.51 | 508 |
| Proposed_II | 91.37 | 705 |

## 4. CONCLUSION

A new splitting method for slimming the CNN structures was proposed. The proposed algorithm split the network structure block by block based on their capability to being split. A fusion block is dedicated after each split block, to preserve global information. In the experimental results, it was observed that splitting a network structure was not possible with a straightforward approach. Also, the proposed algorithm split the VGG16 network structure with an average split factor 4, with only 0.61 loss of accuracy. Finally, it can be noted that splitting the network structure could be used to reduce the number of operations and memory storage requirements effectively.